\documentclass[10pt,journal,compsoc]{IEEEtran}
\usepackage{cite}
\usepackage{amsmath,amssymb,amsfonts}
\usepackage{algorithmic}
\usepackage{graphicx}
\usepackage{textcomp}
\usepackage{comment}
\usepackage{threeparttable}
\usepackage{hyperref}
\usepackage{threeparttable}
\usepackage{multirow}
\usepackage{booktabs}
\usepackage{upgreek}
\def\BibTeX{{\rm B\kern-.05em{\sc i\kern-.025em b}\kern-.08em
    T\kern-.1667em\lower.7ex\hbox{E}\kern-.125emX}}

\begin{document}

\title{Robust Multimodal Learning Framework For Intake Gesture Detection Using Contactless Radar and Wearable IMU Sensors}
\author{
Chunzhuo Wang,  Hans Hallez,  Bart Vanrumste
\thanks{This work is funded in part by the Horizon Europe Research and Innovation Program under Grant GA.No. 101083388, in part by the Flanders AI Research Program, and in part by the Leuven.AI Institute.}
\thanks{Chunzhuo Wang, and Bart Vanrumste are with the e-Media Research Lab, and also with the ESAT-STADIUS Division, KU Leuven, 3000 Leuven, Belgium (e-mail: chunzhuo.wang@kuleuven.be; bart.vanrumste@kuleuven.be).}
\thanks{Hans Hallez is with the M-Group, DistriNet, Department of Computer Science, KU Leuven, 8200 Sint-Michiels, Belgium (e-mail: hans.hallez@kuleuven.be).}
}

\maketitle
\begin{abstract}
	Automated food intake gesture detection is a critical task in the domain of dietary monitoring, with the potential to objectively and continuously record individuals' eating patterns, thereby improving the quality of life (QoL) for diverse populations. Wrist-worn inertial measurement units (IMUs), have been extensively explored for this task and have demonstrated promising performance. Recently, ambient-based contactless radar sensor, has also shown feasibility for intake gesture detection. This study aims to investigate whether the complementary features of wearable and contactless sensors can be effectively leveraged through multimodal learning to further enhance detection performance. Additionally, this study addresses a key challenge in multimodal learning, the reduced robustness when handling missing modalities. To this end, we propose a robust multimodal temporal convolutional network with cross-modal attention (MM-TCN-CMA) framework, designed to efficiently integrate information from IMU and radar sensors, improve detection performance, and maintain effectiveness when modality-incomplete data are encountered during the inference phase. A dataset comprising 52 meal sessions (3,050 eating gestures and 797 drinking gestures) collected from 52 participants is developed and made publicly available for validation. Experimental results demonstrate that the proposed fusion framework achieves a segmental F1-score improvement of 4.3\% and 5.2\% over unimodal-Radar and unimodal-IMU baselines, respectively. Under missing modality conditions, the framework still yields performance gains of 1.3\% and 2.4\% in the missing-Radar and missing-IMU scenarios, respectively. To the best of our knowledge, this is the first study to explore a robust multimodal learning framework combining IMU and radar. The proposed robust radar-IMU fusion framework holds potential for broader applications in other continuous, fine-grained human activity recognition (HAR) tasks.
\end{abstract}


\section{Introduction}
\IEEEPARstart{A}{utomated} intake gesture detection is an important task in food intake monitoring domain, which has drawn lots of attention due to its potential application in personalized nutrition and the intervention of dietary-related health problems \cite{b1,b2}. Beyond its importance in the context of food intake, intake gesture detection also presents a challenging problem within the broader field of human activity recognition, owing to several unique characteristics. First, intake gestures represent continuous activities embedded within real-world scenarios (e.g., a meal session or full-day monitoring), and are often interwoven seamlessly with other daily activities. Second, intake gestures exhibit considerable variability due to differences in utensils, eating habits, and factors such as age or social context (eating alone versus eating in a group). Third, the duration of intake gestures varies substantially across instances. Based on prior studies \cite{b4,10584254}, the intake gesture (eating and drinking) is defined as a series of movements from raising the hand to the mouth with cutleries/liquid container until the hand is moved away from the mouth. 

To enable the automatic and efficient detection of intake gestures, a variety of sensors combined with machine learning techniques have been explored over the past decade. These efforts include wearable-based inertial measurement unit (IMU) \cite{b4,b5,b6}, ambient-based camera\cite{b7}, and portable-based smart food plate \cite{b8} solutions. Recently, our studies \cite{b9,b24} demonstrated that contactless millimeter-wave frequency-modulated continuous wave (FMCW) radar can also be effectively utilized for intake gesture detection in real-world meal scenarios. Along with findings from other studies \cite{b10,b11,b12}, radar sensing has emerged as a promising technique in this domain. With the growing prevalence of diverse sensors in real-life settings (e.g., smart homes), it has become increasingly common for multiple sensors to coexist. Consequently, researchers have begun leveraging multimodal sensor fusion to compensate for the limitations of individual modalities and to enhance detection performance. Examples include combinations such as IMU-flex sensors \cite{b13}, IMU-proximity sensors \cite{b14}, and IMU-camera systems \cite{b15}. 

However, despite the demonstrated potential of radar as a sensing modality for intake gesture detection, the fusion of IMU sensors and contactless radar sensors has yet to be investigated. We argue that fusing radar and IMU represents a promising and practical combination in real-world applications. IMU sensors are already widely integrated into consumer devices such as smartwatches and fitness trackers, while radar sensors are gaining traction in indoor environments due to their advantages of low privacy concerns, independence from lighting conditions, and contactless sensing capabilities \cite{10327733}. These modalities offer complementary sensing perspectives: wrist-worn IMUs provide fine-grained motion and orientation data from an egocentric viewpoint, capturing the nuanced dynamics of hand and arm gestures, whereas FMCW radar sensors enable contactless sensing, capturing global spatial and velocity information without requiring direct attachment to the body, offering an exocentric view.

\begin{figure}[t]
	\centering
	\includegraphics[scale=0.32]{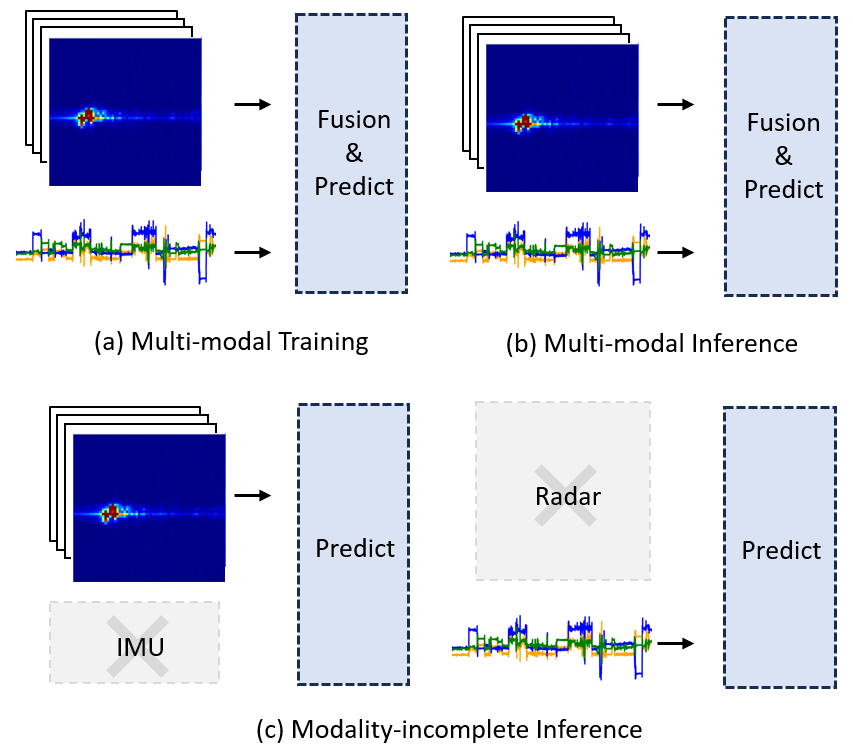}      
	\caption{Training and inference procedures in robust multimodal learning. (a) Training: both radar and IMU modalities are utilized for model training; (b) Inference: both radar and IMU modalities are available during testing; (c) Inference: only the radar modality is available during testing, or only the IMU modality is available during testing.}  
	\label{multi_visual}
\end{figure}

While maximizing the information collected from multiple modalities holds great promise for improving human activity recognition (HAR) performance, a common limitation of multimodal systems is their inefficiency in handling missing modality problems. Specifically, multimodal models trained with complete modalities often experience performance degradation compared to unimodal models when incomplete modality information is encountered during the testing phase \cite{woo2023towards}. Consequently, a more challenging problem arises: how to design multimodal models that are robust and capable of effectively processing both modality-complete and modality-incomplete data during inference, as illustrated in Fig. \ref{multi_visual}.

This challenge is particularly significant in multimodal learning involving contactless radar and wearable IMU sensors. These two modalities are typically deployed in different spatial contexts, wearable devices are attached to the body, while radar sensors are installed in the environment. On one hand, they offer strong complementary sensing capabilities; on the other hand, the likelihood of both modalities being simultaneously available is lower compared to systems which are integrated into a single device. To date, robust multimodal learning methods capable of addressing missing modalities in the HAR domain have not been widely explored.

In this paper, we propose a robust multimodal temporal convolutional network combined with cross-modal attention (MM-TCN-CMA) framework for efficiently fusing radar and IMU sensors to achieve fine-grained intake gesture detection during meal sessions. The main contributions of this research can be summarized as follows:
\begin{itemize}
	\item A multimodal intake gesture detection system that combines radar and IMU sensors has been proposed. The system effectively fuses features from the two modalities and leverages their complementary information, resulting in improved performance for fine-grained intake gesture detection during continuous meal sessions with statistic significance. 
	\item A missing modality handling mechanism has been integrated into the multimodal intake gesture detection approach, enabling the system to maintain robustness when faced with incomplete sensor data during inference. To the best of our knowledge, this is the first study to investigate robust multimodal learning for continuous HAR based on IMU-Radar fusion. Furthermore, we demonstrate that the proposed framework is compatible with multiple models, showing its model-agnostic feature.
	\item A publicly available Radar-IMU multimodal dataset has been collected as part of this study\footnote{We plan to add the link of this dataset in the revision stage.}. The dataset comprises 52 continuous meal sessions collected from 52 participants and represents the first public multimodal HAR dataset that combines wrist-worn IMU data with contactless radar sensing.
\end{itemize}

\section{Related Work}
In this section, we first review existing studies on food intake gesture detection, covering both unimodal and multimodal approaches. We then introduce research efforts focused on multimodal learning in HAR. Subsequently, we discuss specific studies on radar–IMU fusion for HAR tasks. Finally, we introduce robust multimodal learning approaches designed to address missing modality challenges in other domains.

\subsection{Food Intake Gesture Detection}
\subsubsection{Unimodal Approach}
Various sensors have been employed for unimodal intake gesture detection, which can be categorized into three groups based on their deployment position: wearable, portable, and ambient systems. Wearable systems primarily utilize IMU sensors embedded in wristbands \cite{b4,10684446}, small-sized cameras \cite{9187935}, and acoustic sensors \cite{7999175} configured as necklaces for intake gesture detection. Additionally, photoplethysmography (PPG) sensors \cite{7592214} and electromyography (EMG) sensors \cite{7516224} have also been explored for bite detection. Portable solutions include devices such as smart plates \cite{b8} and snacking boxes \cite{DEGOOIJER2023105002} equipped with pressure sensors. In terms of ambient solutions, cameras have been the most common choice \cite{9187935}. More recently, radar has emerged as a promising alternative for intake gesture detection \cite{b9}.

\subsubsection{Multimodal Approach}
Several studies have investigated the fusion of multiple modalities for intake gesture monitoring to further enhance detection performance. Bedri et al. \cite{b14} proposed FitByte, a wearable system consisting of eyeglasses integrated with a camera, a proximity sensor, and six IMUs, detect intake gestures. Wu et al. \cite{b16} developed an ambient-based multimodal system incorporating an RGB camera, a depth camera, and a radar sensor. Notably, these solutions typically integrate multiple sensors into a single device to create either a fully wearable or a fully ambient-based system. Alternatively, Heydarian et al. \cite{b15} explored the fusion of wearable and ambient sensors. Specifically, an IMU-based wristband and an exocentric camera, for intake gesture detection. Their fused approach achieved a higher F1-score compared to using either the inertial or video modality alone. Although these multimodal approaches have demonstrated improvements in intake gesture detection performance, it remains an open question whether fusing wearable-based IMU sensors with ambient-based radar sensors can further enhance performance.

\subsection{Multimodal Fusion in HAR}
\subsubsection{Basic Fusion Technique}
Sensor-based multimodal learning has been extensively studied across various tasks, including daily activity recognition \cite{s22010174, 10529207} and gait analysis \cite{Yang2024315880}. Three primary approaches are commonly employed for fusing data from different modalities: data-level fusion, feature-level fusion, and decision-level fusion. In data-level fusion, raw data from multiple sensors are concatenated along the feature dimension to form a unified input for subsequent processing \cite{Yang2024315880}. In feature-level fusion, sensor data are first processed, either through deep learning models or traditional feature engineering methods, to extract higher-level representations, which are then combined and used for prediction \cite{s22010174, 10529207}. In decision-level fusion, data from each sensor are independently processed to generate separate predictions, which are subsequently processed (e.g., through majority voting or averaging) to produce a final decision.

\subsubsection{Radar-IMU Fusion in HAR}
Although no existing studies have specifically explored IMU–Radar fusion for intake gesture detection, two studies have investigated the integration of IMU and radar sensors in HAR. Li et al. \cite{b19} proposed a hybrid fusion approach based on Bi-LSTM networks to classify human activities, including fall detection, using IMU and FMCW radar data. Their experiment involved 16 participants and generated 288 recorded events. Recently, Yu et al. \cite{b20} combined data from IMU, radar, and a universal software radio peripheral (USRP) to classify two activities: sit-down and stand-up, using manually selected features and a discrete Hopfield neural network (DHNN). While these approaches demonstrated promising performance using hand-crafted features, end-to-end learning methods without manual feature engineering have yet to be thoroughly explored. Moreover, both studies focused on controlled environments with scripted data collection protocols. As emphasized in our previous work, intake gesture detection during meal sessions represents a form of fine-grained, continuous gesture detection. Therefore, this study aims to investigate the potential of Radar-IMU fusion for detecting more realistic and challenging continuous activities derived from real-world scenarios.

\subsubsection{Robust Multimodal Learning with Missing Modality}
Given the inherent difficulty of obtaining modality-complete data consistently, the design of robust multimodal learning models capable of handling missing modalities has emerged as a critical area of research. Existing studies demonstrate that standard multimodal models experience a sharp decline in performance when confronted with incomplete modality data, often performing worse than unimodal models \cite{woo2023towards}. Current research primarily focuses on modality incompleteness across various tasks, including semantic understanding \cite{9879085, woo2023towards}, medical imaging segmentation \cite{10184044}, emotion recognition \cite{zuo2023exploiting}, and sleep staging \cite{10400520}. These solutions can be categorized into two main approaches: 1) missing data reconstruction and 2) joint representation learning. The missing data reconstruction approach aims to recover the missing modality based on the available modalities, often utilizing generative models or autoencoders. However, the overall performance is heavily dependent on the quality of the reconstructed modalities. On the other hand, joint representation learning seeks to extract shared latent features from either fully observed multimodal data or a subset of available modalities. The central premise of this approach is that different modalities share common features. While this method shows promise, it is generally more effective when applied to datasets containing more than two modalities and typically requires larger datasets for effective training \cite{zuo2023exploiting,ZHOU2023109665}.

In contrast to the aforementioned tasks, the study of robust multimodal learning for sensor-based HAR remains relatively underexplored. In the Centaur study \cite{10506095}, the authors proposed a convolutional layer-based autoencoder to reconstruct missing data, followed by convolutional and attention layers for multimodal fusion and activity detection. The modalities investigated included IMU data (collected from smartwatches and smartphones) and Electrocardiogram (ECG) signals (acquired from a chest sensor). However, this study primarily addresses scenarios involving partial data loss over short durations relative to the entire session. To date, no HAR studies have specifically tackled situations where data is missing for an entire session, such as a complete meal session. Furthermore, prior research typically assumes multimodal data with similar or identical dimensionalities, leaving a gap in addressing the challenges posed by multimodal data of heterogeneous dimensions. In this study, we propose a feature reconstruction-based approach to overcome the limitations identified in previous HAR research and multimodal food intake gesture detection studies. To the best of our knowledge, no existing work has developed a robust multimodal learning framework for sensor-based HAR tasks that leverages feature reconstruction, particularly by incorporating both Radar and IMU data.

\section{Data Collection and Preprocessing}
\subsection{Sensors}

\begin{figure}[t]
	\centering
	\includegraphics[scale=0.30]{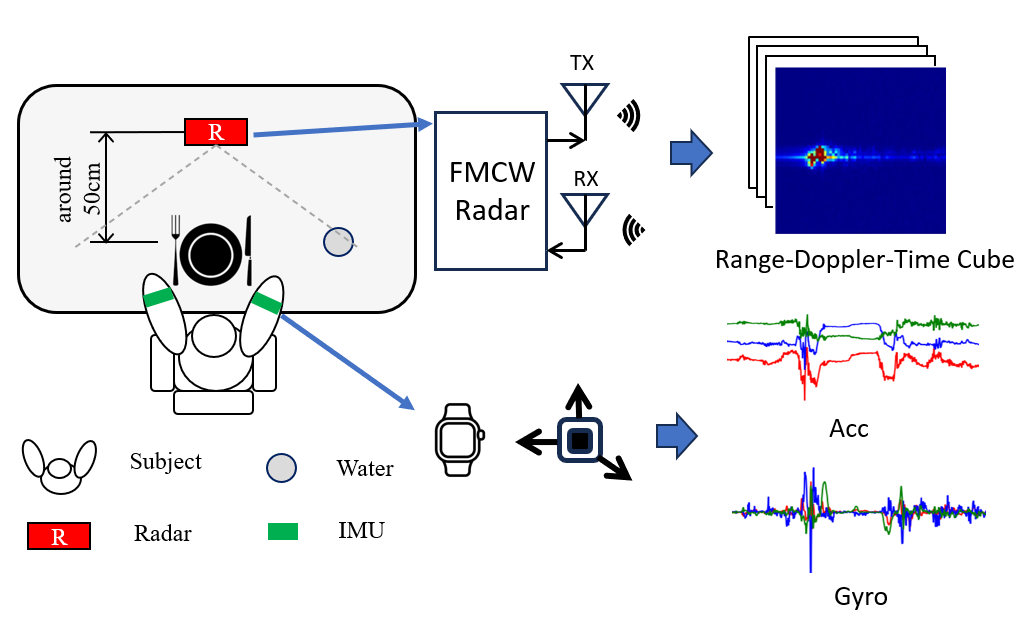}      
	\caption{Data experiment scene.}  
	\label{exp_setup}
\end{figure}

An FMCW 76GHz IWR1843 radar from Texas Instrument (TI)\footnote{https://www.ti.com/tool/IWR1843BOOST} and two Shimmer3 IMU sensors\footnote{https://shimmersensing.com/product/shimmer3-imu-unit/} were utilized for data collection. The radar system parameters were configured as described in \cite{b9}. The radar generates 25 frames of raw data per second (fps), with each frame having a dimension of $128\times128$. Two IMU sensors were attached to the participants' wrists, each operating at a sampling rate of 64 Hz. Each IMU sensor contains a 3-axis accelerometer and a 3-axis gyroscope, providing six degrees of freedom (DoF). In addition to the radar and IMU sensors, a camera was deployed to record the experimental process for annotation and temporal alignment purposes. The experimental setup is depicted in Fig.~\ref{exp_setup}.

\subsection{Data Collection and Annotation}
This study was approved by the KU Leuven ethical committee (Reference Number: G-2021-4025-R4), and written informed consent was obtained from all participants prior to data collection. A total of 52 participants (17 females and 35 males) were recruited, each participating in a single meal session. Data collection was conducted across seven different rooms with varying layouts at the KU Leuven University Campus Group T, including one staff office, three classrooms, two meeting rooms, and one study room. The total duration of the collected dataset amounts to 894 minutes, with an average meal duration of 17.19 $\pm$ 5.44 minutes. During the experiment, participants were seated directly in front of the radar while eating, with the distance between the radar and the food plate ranging from 0.3 m to 0.6 m, as illustrated in Fig. \ref{exp_setup}. Detailed information regarding the meals is provided in Table \ref{meal_data}. It is noteworthy that a portion of the data was collected concurrently with our previous Eat-Radar project \cite{b9}. Participants were instructed to eat in their natural manner, which inherently included non-feeding movements during the meal sessions, such as touching their nose, combing their hair with their hands, adjusting their eyeglasses, using smartphones.

The ELAN tool \cite{b14} was utilized to annotate the video recordings and to align the timestamps of the radar and IMU signals. To enable accurate synchronization, participants were instructed to perform a quick up-and-down motion with both hands as a motion indicator before starting their meal. Video recordings served as the reference for aligning IMU and radar data using this motion cue. Once each modality was aligned with the video, the IMU and radar signals were inherently synchronized. The annotated data were categorized into three classes: eating, drinking, and other activities. It is important to note that food intake gestures (both eating and drinking) exhibit substantial variability between participants, and even within a single meal session for the same individual. Therefore, the endpoint of a food intake gesture was defined as the moment the hand moves away from the mouth, rather than strictly when it returns to the plate or table. In total, 3,050 eating gestures and 797 drinking gestures were recorded. Detailed statistics of the dataset are summarized in Table~\ref{eat_drink_data}.

\begin{table}[t]
	\caption{Meal data statistics}
	\label{meal_data}
	
	\begin{center}
		\scalebox{1}{
			\begin{tabular}{lr}
				\toprule
				\hline
				Parameter & Values\\
				\midrule
				Participants  &52 \\
				Female : male & 17:35 \\
				Duration ratio of other : eating : drinking& 9.44:2.21:1\\
				Mean meal duration &(17.19$\pm$5.44) min \\
				Total duration & 894 min\\
				\hline
				\bottomrule
		\end{tabular}}
	\end{center}
\end{table} 

\begin{table}[t]
	\caption{Eating and Drinking gesture Details}
	\label{eat_drink_data}
	\begin{center}
		\scalebox{1}{
			\begin{tabular}{lcc}
				\toprule
				\hline
				\multirow{2}*{Parameter} & \multicolumn{2}{c}{Values}\\
				~& Eating & Drinking\\
				\midrule
				Number of gestures& 3,050& 797\\
				
				Duration range of gestures & 1.00-17.08 s& 2.36-24.56 s\\
				Median duration of gestures & 2.72 s& 4.68 s\\
				Mean $\pm$ std of gestures & 3.07 $\pm$ 1.42 s& 5.32 $\pm$ 2.42 s\\
				\hline
				\bottomrule
		\end{tabular}}
	\end{center}
\end{table} 

\begin{figure*}[t]
	\centering
	\includegraphics[scale=0.60]{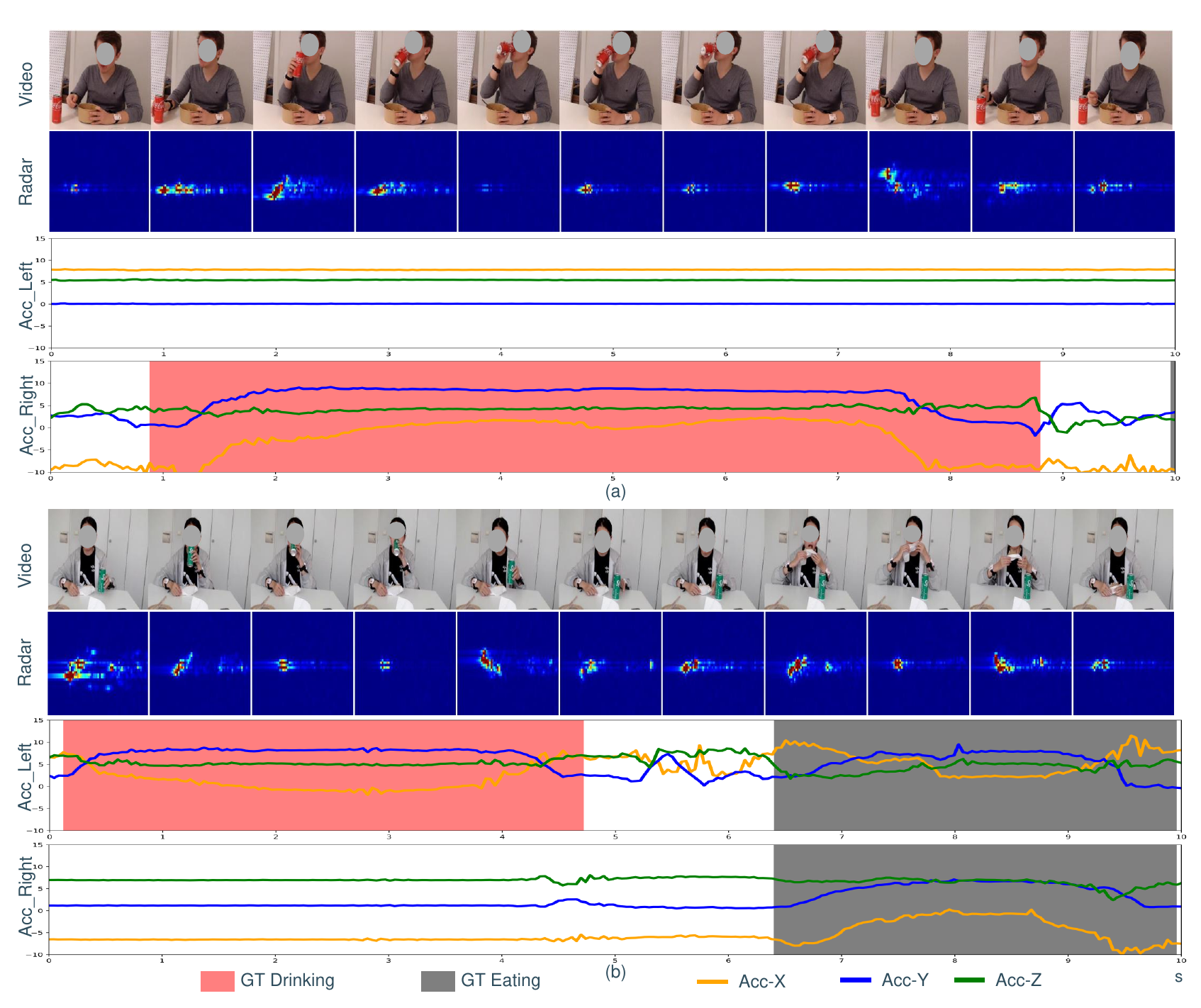}      
	\caption{Examples of radar range-doppler-time data and 3-axis accelerometer data from dual-hand IMU sensors for intake gestures. In (a), the participant uses the right hand to drink while the left hand remains static. In (b), the participant initially uses the left hand to drink, followed by a bimanual action to eat a sandwich.}  
	\label{data_repre}
\end{figure*}

\subsection{Data preprocessing}
\subsubsection{Radar Data Preprocessing}
The range-Doppler-time (RDT) cube was used as the input representation for the radar modality. During each meal session, the radar continuously generated raw data at 25 fps. Several preprocessing steps were applied to the raw data, including Fast Fourier Transform (FFT) for range measurement (Range-FFT), static object removal, FFT for velocity measurement (Doppler-FFT), superposition, and cropping, following the procedure described in \cite{b9}. After preprocessing, the RDT cube was structured with dimensions $N_r \times N_d \times N = 32 \times 64 \times N$, where $N_r$ denotes the number of range bins (maximum range: 1.28 m), $N_d$ the number of Doppler bins (maximum velocity: $\pm$ 1.28 m/s), and $N$ the number of frames in a meal session. 
\subsubsection{IMU Data Preprocessing}
Two wrist-worn IMU sensors were utilized in this study, each providing 6 DoF data. The IMU signals from both hands were concatenated along the feature channel, resulting in a 12-dimensional feature set. The IMU data for one meal session can thus be represented as a matrix of size $N_i \times N = 12 \times N$, where $N_i$ denotes the number of features and $N$ the number of sample points in the session. To achieve sample-wise alignment between the radar and IMU signals, the IMU data were downsampled from 64 Hz to 25 Hz. Our previous studies in \cite{b25} have indicated that IMU signals sampled above 16 Hz do not substantially impact model performance. Examples of RD frames and IMU signals aligned with eating movements are illustrated in Fig.~\ref{data_repre}.

\section{Robust Multimodal Framework}
\subsection{Problem Definition}
To simplify the problem without loss of generality, we consider a modality-complete dataset comprising two modalities: the RDT Cube ($x_{R}$) and IMU data ($x_{I}$). Each sample can be denoted as ${x_{R}, x_{I}, y}$, where $y$ represents the corresponding label. Similarly, modality-incomplete datasets are represented as ${x_{R}, y}$ (missing IMU data) or ${x_{I}, y}$ (missing radar data). The objective is to leverage the modality-complete dataset to train a multimodal learning model capable of efficiently processing both modality-complete and modality-incomplete data during the inference phase, with an emphasis on achieving higher performance compared to unimodal approaches. 

\begin{figure*}[t]
	\centering
	\includegraphics[scale=0.3]{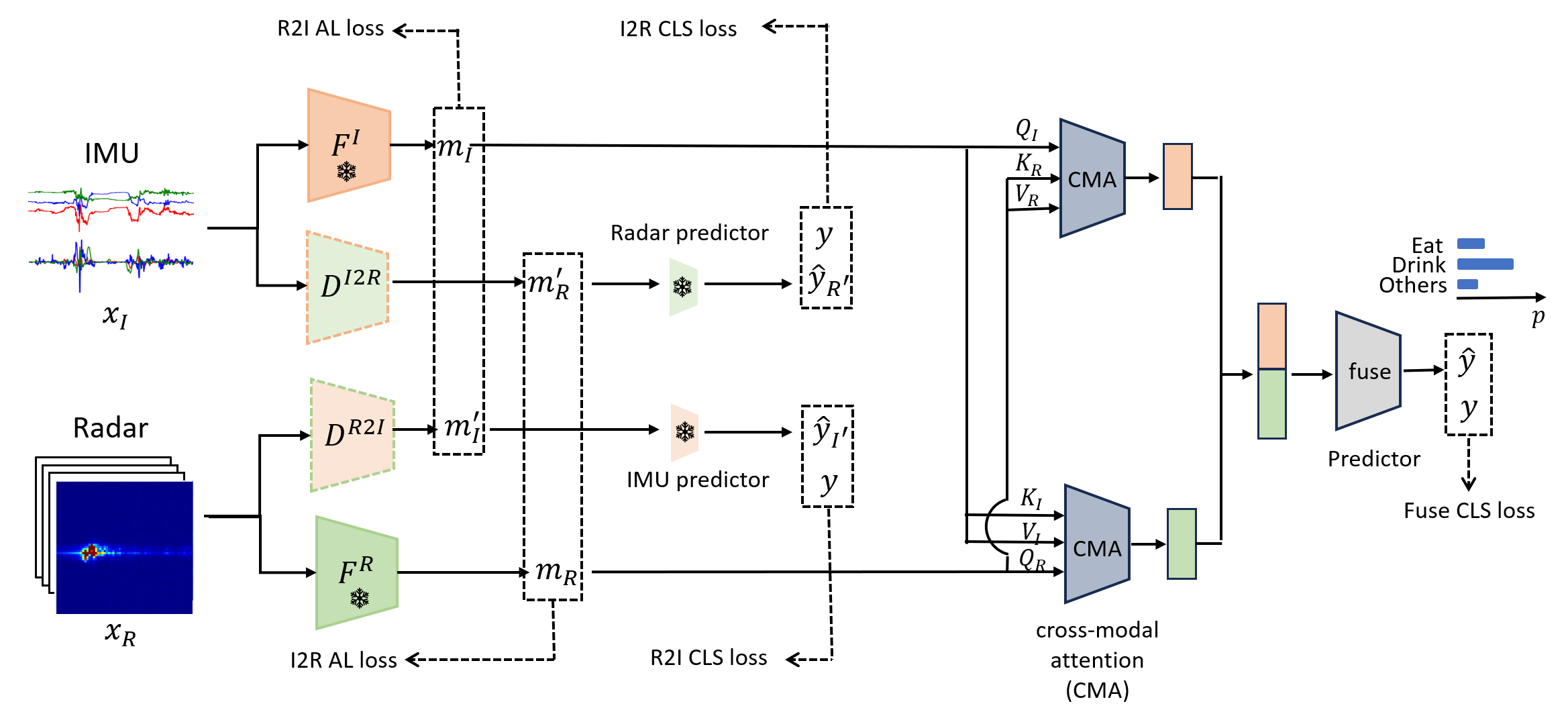}      
	\caption{Overall architecture of the proposed robust multimodal learning framework. Input data from the IMU and Radar modalities, denoted as $x_I$ and $x_R$, are processed by the modality specific feature extraction encoders (MSFE), $\mathbb{F}^I$ and $\mathbb{F}^R$, to obtain modality-specific intermediate features $m_I$ and $m_R$. Simultaneously, $x_I$ and $x_R$ are fed into the modality adaption encoders (MAE), $\mathbb{D}^{I2R}$ and $\mathbb{D}^{R2I}$, to generate adapted features $m'_R$ and $m'_I$. The cross-modal attention (CMA) modules are then employed to fuse the extracted features. The snow symbol represents that the module with frozen weights.}
	\label{model_overview}
\end{figure*}

\subsection{Model Architecture}
The proposed robust multimodal learning framework comprises four key components: modality-specific feature extraction encoders (MSFE) for extracting representative features from each modality, modality adaptation encoders (MAE) for generating target modality features from source modality data, a feature fusion module (FFM) for integrating multimodal information, and a final classifier for prediction (Predictor).
\subsubsection{Modality-specific Feature Extraction (MSFE)}
The feature extraction encoders are designed to extract modality-specific features from each input modality. For the IMU modality, the encoder is denoted as $\mathbb{F}^I: x_I \rightarrow m_I$, where $x_I \in \mathbb{R}^{12\times{N}}$ represents the input IMU data, and $m_I \in \mathbb{R}^{64\times{N}}$ denotes the extracted intermediate IMU features. The radar feature extraction encoder is denoted as $\mathbb{F}^R: x_R \rightarrow m_R$, where $x_R \in \mathbb{R}^{32\times{64}\times{N}}$ corresponds to the input RDT radar data, and $m_R \in \mathbb{R}^{64\times{N}}$ represents the extracted intermediate radar features.  

\subsubsection{Modality Adaption Encoders (MAE)}
To address the issue of missing modalities, inspired by the IMU2Doppler framework proposed by Bhalla et al. \cite{b21}, we designed a modality adaptation encoder (MAE) to reconstruct the features of a missing modality using information from the available modality. The module responsible for generating radar features from IMU data (IMU-to-Radar) is denoted as $\mathbb{D}^{I2R}: x_I \rightarrow m'_R$, where $m'_R \in \mathbb{R}^{64\times{N}}$ represents the reconstructed intermediate radar features. Conversely, for reconstructing IMU features from radar data (Radar-to-IMU), the corresponding module is denoted as $\mathbb{D}^{R2I}: x_R \rightarrow m'_I$, where $m'_I \in \mathbb{R}^{64\times{N}}$ represents the reconstructed intermediate IMU features.

\subsubsection{Feature Fusion Module (FFM)}
The extracted intermediate features $m_R$ and $m_I$ are fused through a feature fusion module (FFM). To implement the FFM, we investigated four fusion techniques: (1) element-wise addition, (2) channel-wise concatenation, (3) decision-level fusion, and (4) cross-modal attention (CMA) \cite{woo2023towards}:

1) Element-wise addition: This method directly combines the features by adding $m_R$ and $m_I$ element-wise: \begin{equation} m_{\text{fused}} = m_R + m_I \end{equation}

2) Channel-wise concatenation: The intermediate features $m_R$ and $m_I$ are concatenated along the channel dimension: \begin{equation} m_{\text{fused}} = \text{concat}(m_R, m_I) \end{equation}

3) Decision-level fusion: Here, the prediction probabilities from the IMU and radar modalities are first computed separately. These probabilities are then summed and averaged to obtain the final fused prediction.

4) Cross-modal attention (CMA): The CMA, derived from self-attention mechanisms, is utilized to integrate complementary information between modalities. Unlike self-attention, where the queries (Q), keys (K), and values (V) are from the same modality, CMA uses queries from one modality and keys and values from the other. Specifically, the cross-attention from IMU to radar is formulated as:  
\begin{equation}
	\text{crs-att}_{I \rightarrow R} = \text{softmax}\left(\frac{Q_RK_I^T}{\sqrt{d_k}}\right)V_I
\end{equation}

To fully leverage information from both modalities, a symmetric cross-modal attention mechanism is employed, where features are fused bidirectionally:
\begin{equation}
	m_{\text{fused}} = \text{concat}(\text{crs-att}_{I \rightarrow R}, \text{crs-att}_{R \rightarrow I})\label{eq4}
\end{equation}

Based on experimental results, the CMA method was ultimately selected for feature fusion. A detailed discussion and comparison of the different fusion strategies are provided in Section \ref{sec: abla}.

\subsection{Loss Function} 
\subsubsection{Sample-wise Classification Loss Function}
\label{sec:clas}
A classification loss and a smoothing loss are combined to form the sample-wise loss function. First, a cross-entropy loss is used to represent the classification loss:
\begin{equation}\mathcal{L}_{ce}=\frac{1}{N}\sum_{t,c}-{y_{t,c}}{\rm log}(\hat{y}_{t,c})\label{eq4}\end{equation}

where $y_{t,c}$ is the ground truth label, and $\hat{y}_{t,c}$ is the predicted probability for class $c$ at time $t$.

Second, to further improve prediction quality, a truncated mean squared error (T-MSE) over sample-wise log-probabilities is applied as a smoothing loss:
\begin{equation}\mathcal{L}_{T-MSE}=\frac{1}{NC}\sum_{t,c}\widetilde{\triangle}{^2}_{t,c}\label{eq5}\end{equation}
\begin{equation}\widetilde{\triangle}_{t,c}=\begin{cases}
		\triangle_{t,c} : \triangle_{t,c}\leq{\tau}\\
		\tau : otherwise\\
	\end{cases}
	\label{eq6}\end{equation}
\begin{equation}\triangle_{t,c} =|{\rm log}(\hat{y}_{t,c})-{\rm log}(\hat{y}_{t-1,c})|\label{eq7}\end{equation}
where $N$ is the data length, $C$ is the number of classes, $\hat{y}_{t,c}$ is the probability of class $c$ at time $t$, and $\tau$ is a threshold to truncate the mean squared error. The smoothing function is used to reduce the over-segmentation errors by calculating the class log probability deviation between adjacent prediction points \cite{b37}. The loss function is obtained to combine the mentioned losses :
\begin{equation}\mathcal{L}_{cls-fuse} =\mathcal{L}_{ce}+\lambda{\mathcal{L}_{T-MSE}}\label{eq8}\end{equation}
where $\lambda$ is a parameter to determine the contribution of the two losses. The $\tau$ and $\lambda$ are set to 4 and 0.15 from \cite{b37}.

\subsubsection{Modality Adaption Loss Function}
The feature alignment loss function is employed to encourage the MAE to generate intermediate features from the source modality that closely match the features extracted by the MSFE of the target modality. For MAE $\mathbb{D}^{R2I}$, the mean square error (MSE) is adopted as the alignment loss, defined as follows:

\begin{equation}
	\mathcal{L}_{al-R2I} = \frac{1}{n} \|  m'_I - m_I \|_2^2
	\label{eq8}\end{equation}

where $\mathcal{L}_{al-R2I}$ denotes the alignment loss between the original IMU features $m_I$ from MSFE $\mathbb{F}^I$ and the reconstructed IMU feature $m'_I$ from MAE $\mathbb{D}^{R2I}$, $n$ denotes the total number of elements $64\times{N}$. To further enhance the classification performance of the MAE, a classification loss, denoted as $\mathcal{L}_{cls-R2I}$, is also applied to the reconstructed features $m'_I$ using an IMU predictor. The classification loss is consistent with the formulation described in Section~\ref{sec:clas}. Consequently, the total modality adaptation loss for the R2I direction is defined as the sum of the feature alignment loss and the classification loss:

\begin{equation}\mathcal{L}_{R2I} = \mathcal{L}_{al-R2I}+\beta\mathcal{L}_{cls-R2I}\label{eq8}\end{equation}

where $\beta$ is a parameter to determine the contribution of the two losses and is set to 0.35 according to \cite{b21}. Similarly, $\mathcal{L}_{\text{I2R}}$ is defined for the reverse direction, adapting IMU features to radar features.
\subsubsection{Overall Loss Function}
The overall loss function combines the primary classification loss and the auxiliary modality adaptation losses, and is formulated as:
\begin{equation}
	\mathcal{L}_{\text{total}} = \mathcal{L}_{\text{cls-fuse}}+\mathcal{L}_{\text{R2I}}+\mathcal{L}_{\text{I2R}}
\end{equation}

\subsection{Training and Testing}  
\label{sec:tran}
A two-step training strategy is employed during the training phase. In the first step, unimodal models for the IMU and radar modalities are independently trained. Each unimodal model consists of two components: a MSFE block to extract intermediate representations, and a Predictor to generate class probabilities from these intermediate features. In the second step, the trained MSFEs and Predictors, with their parameters frozen, are integrated into the overall multimodal learning framework. Subsequently, the remaining modules, such as the MAE, CMA, and multimodal predictor, are trained within this framework, as illustrated in Fig. \ref{model_overview}. This strategy ensures that $\mathbb{D}^{I2R}$ learns to mimic $\mathbb{F}^I$ in a one-way manner, thereby preventing the undesirable scenario where $\mathbb{F}^I$ adapts to $\mathbb{D}^{I2R}$, which could result in suboptimal performance. It is important to note that the Radar Predictor and IMU Predictor are utilized exclusively during the training phase.

During the testing phase, under a modality-complete scenario, the outputs of $\mathbb{F}^I$ and $\mathbb{F}^R$, denoted as $m_I$ and $m_R$, respectively, are input into the cross-modal attention module for feature fusion. In modality-incomplete scenarios, such as when the IMU modality is missing, the outputs of $\mathbb{D}^{R2I}$ and $\mathbb{F}^R$, denoted as $m'_I$ and $m_R$, respectively, are used as inputs to the cross-modal attention module for feature fusion.

\subsection{Implementation}
Building upon our previous study \cite{10584254}, a 1D-TCN is employed as the modality-specific feature extractor (MSFE) for the Uni-IMU branch. For the radar modality, a 3D-TCN, as introduced in \cite{b9}, is adopted as the MSFE for the Uni-Radar branch. To simplify the overall framework complexity, the MAE module for each modality shares the same architecture as the corresponding MSFE, e.g., $\mathbb{D}^{I2R}$ and $\mathbb{F}^{I}$. In the CMA module, eight attention heads are utilized, with each head having a dimensionality of 8. Model training is performed using the Adam optimizer with a learning rate of 0.0005. The number of training epochs is set to 100. The input window length is configured to 40 seconds, and the batch size is set to 4. All training, validation, and testing experiments were conducted using two NVIDIA Tesla A100-SXM2-80GB GPUs provided by the Vlaams Supercomputer Centrum (VSC) \footnote{See https://www.vscentrum.be/}.

\section{Evaluation and Validation Scheme}

\subsection{Evaluation Scheme and Index}
The proposed seq2seq model generates multi-class predictions on a sample-wise basis. Since bite-related datasets tend to be imbalanced, we opt to use Cohen's Kappa index to assess the performance of sample-wise classifications. 

\begin{figure}[t]
	\centering
	\includegraphics[scale=0.5]{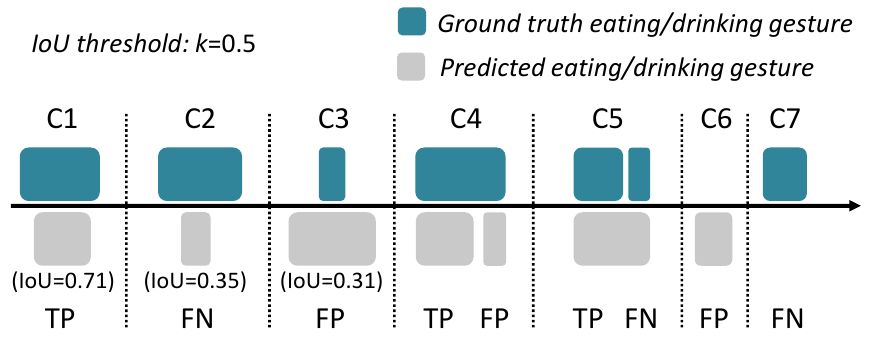}
	\caption{Segment-wise evaluation examples. If the target is drinking class, eating gestures are categorized as other, and vice versa.} 
	\label{seg_eva}
\end{figure}

While sample-wise results provide insight into the model's direct classification performance, it's important to recognize that the goal of bite detection is to accurately count bites, which sample-wise metrics do not directly reflect. To address this limitation, we apply a segment-wise evaluation method, previously employed in related work \cite{b9,10402122}, to assess bite detection. Examples of this evaluation are shown in Fig. \ref{seg_eva}. This approach involves two main steps. First, we compute the intersection over union (IoU) between each predicted bite and its corresponding ground truth bite, as illustrated in Fig. \ref{seg_eva} C1-C3. Then, the calculated IoU is compared against a predefined threshold, $k$, to categorize segment-wise true positives (TP), false negatives (FN), and false positives (FP). Based on these values, we calculate the segmental F1-score for each class (eating and drinking):

\begin{equation}
	F1_{c=\left\{E, D\right\}}=\frac{2TP_c}{2TP_c+FP_c+FN_c}
	\label{eq14}
\end{equation}
where $c$ represents the class, $E$, $D$ represents eating and drinking class, respectively.

The segment-wise evaluation accommodates minor temporal misalignments between the ground truth and predictions, which may arise due to variability in annotation. It also provides clear information on the number of bites detected. By varying the threshold $k$, we are able to assess not only detection accuracy but also segmentation performance. In our evaluation, we used two threshold values, 0.1 and 0.5. Detailed information on this segment-wise evaluation has been introduced in \cite{10596709}.

\subsection{Models for Comparison}
\label{mo_com}
To further evaluate the effectiveness of the proposed framework, various models from existing related studies were selected as backbone modules for the MSFE and MAE components. It is important to note that, to the best of our knowledge, no existing work has addressed the exact same scenario investigated in this study. Therefore, directly comparable benchmark models are not available. Instead, several representative models commonly used in related research areas were adopted for comparative analysis.
\subsubsection{CNN-(Bi)LSTM}
The Convolutional Neural Network with Long Short-Term Memory Network (CNN-LSTM) in MSFE-IMU module comprises three 1D convolutional (1D-CNN) layers followed by a single LSTM layer. The CNN-BiLSTM combines the CNN with a bidirectional LSTM layer (BiLSTM). The CNN section includes three layers, each with 64 kernels of size $3$, and utilizes the ReLU activation function. The output from the CNN block is flattened and subsequently fed into the LSTM layer, which contains 128 hidden units. The MSFE-Radar module is designed with four 3D-CNN layers, configured with 16, 32, 32, and 32 kernels, respectively, each employing a kernel size of $3\times{3}\times{3}$. ReLU activation is applied after each convolutional layer. Additionally, each 3D-CNN layer is followed by a max-pooling layer with a pooling size of $2\times{2}\times{1}$.
\subsubsection{ResNet-(Bi)LSTM}
The ResNet-LSTM architecture within the MSFE-IMU module integrates a one-dimensional Residual Network (1D-ResNet-10) backbone with a subsequent LSTM layer, as described in \cite{b7}. ReLU activation is employed after each convolutional layer. The output from the ResNet backbone is flattened and subsequently input into the LSTM layer, which contains 128 hidden units. The MSFE-Radar module, a 3D-ResNet-18 backbone is utilized, following the design proposed in \cite{hara2018can}.

\begin{table}[b]
	\caption{Sample-wise kappa score results using single modality and multiple modalities}
	\label{sample_results}
	\begin{center}
		\scalebox{1}{
			\begin{threeparttable} 
				\begin{tabular}{l|ccc}
					\toprule
					\hline
					Model&Uni-IMU&Uni-Radar&IMU+Radar\\
					\midrule
					CNN-LSTM& 0.672 & 0.574 & 0.701 \\
					CNN-BiLSTM & 0.718 & 0.651 & 0.750 \\
					ResNet-LSTM & 0.666 & 0.645 & 0.715 \\
					ResNet-BiLSTM & 0.742 & 0.637 & 0.725 \\
					MM-TCN-CMA  & 0.747 & 0.729 & 0.777 \\ 
					\hline
					\bottomrule
				\end{tabular} 
			\end{threeparttable} 
		}
	\end{center}
\end{table}

\begin{table*}[t]
	\caption{Segment-wise F1-scores results using single modality and multiple modalities}
	\label{multi_results_2}
	\begin{center}
		\scalebox{1}{
			\begin{threeparttable} 
				\begin{tabular}{l|c|cc|cc|cc}
					\toprule
					\hline
					\multirow{2}*{Model}&\multirow{2}*{\textit{k}}&\multicolumn{2}{c|}{Uni-IMU}&\multicolumn{2}{c|}{Uni-Radar}&\multicolumn{2}{c}{IMU+Radar}\\
					~&~&{Eating}&{Drinking}&{Eating}&{Drinking}&{Eating}&{Drinking}\\
					\midrule
					CNN-LSTM& \multirow{6}*{0.1}&0.833& 0.874&0.797 & 0.771 & 0.869 & 0.883 \\
					CNN-BiLSTM&~&0.852 &0.870 &0.837 & 0.853 & 0.891 & 0.910 \\
					ResNet-LSTM&~&0.824 &0.834 &0.839 & 0.838 & 0.875 & 0.890 \\
					ResNet-BiLSTM&~&0.868 &0.906 &0.837 & 0.862 & 0.886 & 0.910 \\
					MM-TCN-CMA &~&0.880 & 0.893&0.892 & 0.880 & 0.922 & 0.920 \\
					
					\midrule
					CNN-LSTM& \multirow{6}*{0.5}&0.764& 0.816&0.659 & 0.655 & 0.803 & 0.821 \\
					CNN-BiLSTM&~&0.793 &0.821 &0.735 & 0.782 & 0.833 & 0.863 \\
					ResNet-LSTM&~&0.748 &0.782 &0.717 & 0.752 & 0.796 & 0.836 \\
					ResNet-BiLSTM&~&0.812 &0.861 &0.707 & 0.783 & 0.816 & 0.854 \\
					MM-TCN-CMA &~&0.832 & 0.862&0.841 & 0.838 & 0.884 & 0.887 \\
					\hline 
					\bottomrule
				\end{tabular} 
			\end{threeparttable} 
		}
	\end{center}
\end{table*}

\begin{table}[t]
	\caption{Sample-wise kappa score results on missing modalities}
	\label{sample_missing_results}
	\begin{center}
		\scalebox{1}{
			\begin{threeparttable} 
				\begin{tabular}{l|cc}
					\toprule
					\hline
					Model&Missing Radar&Missing IMU\\
					\midrule
					CNN-LSTM  & 0.702 & 0.598\\
					CNN-BiLSTM  & 0.743 & 0.681 \\
					ResNet-LSTM  &  0.701& 0.655 \\
					ResNet-BiLSTM  & 0.753& 0.642  \\
					MM-TCN-CMA   & 0.761& 0.745  \\
					\hline
					\bottomrule
				\end{tabular} 
			\end{threeparttable} 
		}
	\end{center}
\end{table}

\begin{table}[t]
	\caption{Segment-wsie F1-scores results using MM-TCN-CMA model trained on two modalities, but tested on single modality}
	\label{missing_dam}
	\begin{center}
		\scalebox{0.8}{
			\begin{tabular}{l|cc|cc}
				\toprule
				\hline
				\multirow{2}*{Approach}&\multicolumn{2}{c|}{$k=0.1$}&\multicolumn{2}{c}{$k=0.5$}\\
				~&{Eating}&{Drinking}&{Eating}&{Drinking}\\
				\midrule
				Uni-IMU& 0.880 & 0.893 & 0.832 & 0.862 \\
				Uni-Radar & 0.892& 0.880 & 0.841 & 0.838  \\
				Fusion & 0.922& 0.920 & 0.884 & 0.887\\
				Fusion-missing IMU & 0.906& 0.907 & 0.865 & 0.879 \\
				Fusion-missing Radar & 0.886& 0.920 & 0.845 & 0.894\\
				\hline
				\bottomrule
		\end{tabular}}
	\end{center}
\end{table}

\subsection{Validation Strategy}
To evaluate the model's performance on unseen data, a 5-fold cross-validation strategy is employed. In each fold, 42 out of 52 meals are assigned to the training set, while the remaining 10 meals are used as the test set (with 11 meals allocated to the test set in the fourth and fifth folds). This design ensures comprehensive evaluation across the entire dataset throughout the 5-fold experiments. The specific meal\_id lists corresponding to each fold are provided along with the dataset.

\section{Experimental Results}

\begin{figure}[t]
	\centering
	\includegraphics[scale=0.26]{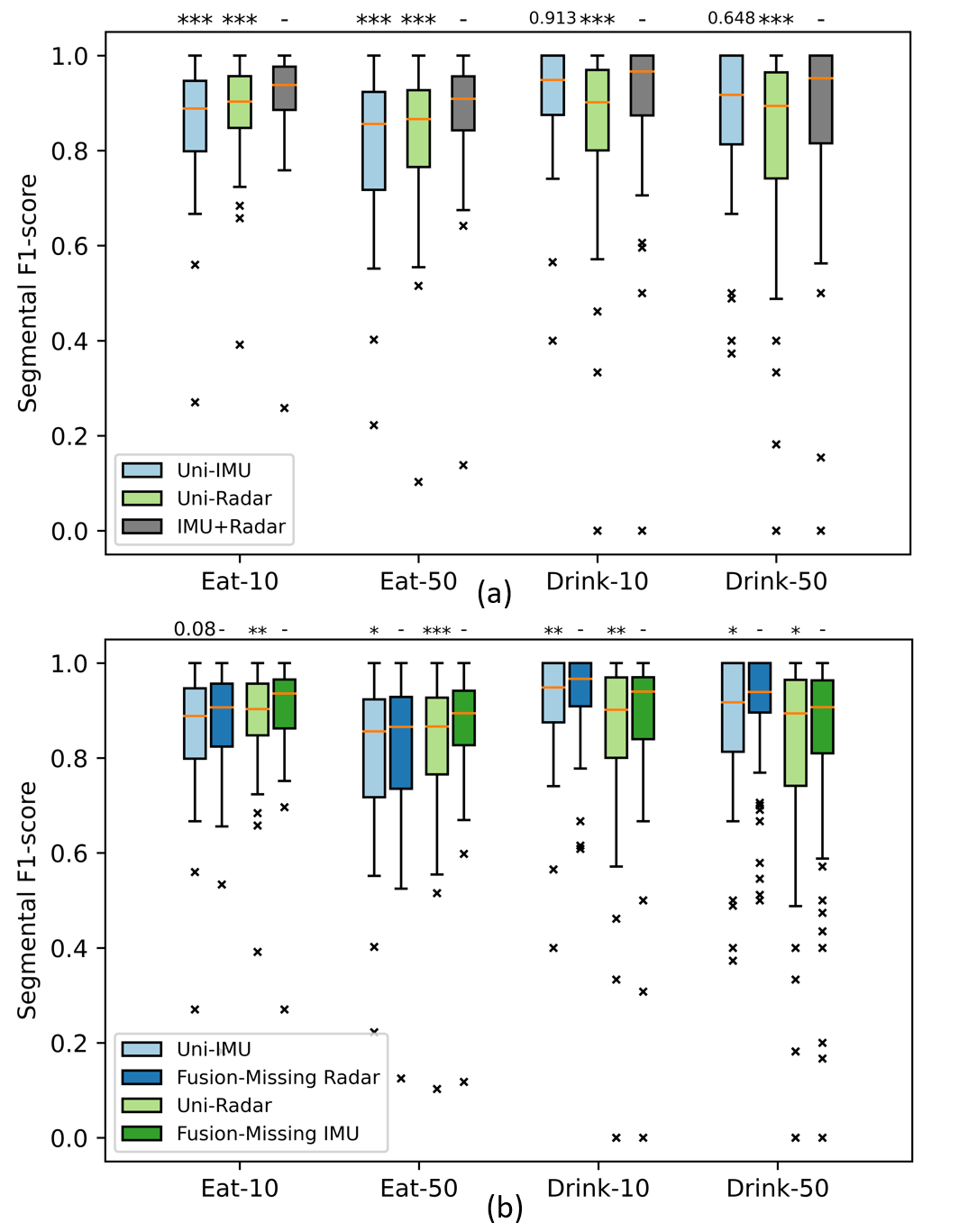}      
	\caption{Boxplots of individual intake gesture detection performance. (a) Performance of unimodal IMU-based, unimodal radar-based, and multimodal fusion approaches using the MM-TCN-CMA model. (b) Performance under missing-modality conditions (missing radar and missing IMU). The y-axis represent the segmental F1-score. In x-axis, Eat-10 and Eat-50 represent eating gesture performance at $k=0.1$ and $k=0.5$, and similar for Drink-10 and Drink-50. The $p$ values are shown on top of the box plots. $p<0.05$: *,$p<0.01$: **,$p<0.001$: ***.}  
	\label{indv_box}
\end{figure}

\subsection{Modality-complete Experiments}
We first conducted experiments to validate the efficacy of the fusion of radar and imu data for intake gesture detection on both sample-wise and segment-wise. In this experiment, different models are used as MSFE module, as mentioned in Section \ref{mo_com}. The sample-wise results are shown in Table \ref{sample_results}. The improved kappa scores have been observed from all models when fusing imu and radar data, with at least 2.8\% improvement except ResNet-BiLSTM model. The MM-TCN-CMA based approach obtained highest kappa score of 0.777, followed by the CNN-BilSM model with kappa score 0.750.  

In the view of segment-wise evaluation scheme, results in Table \ref{multi_results_2} show that the fusion of imu and radar sensors obtained higher eating and drinking gesture detection performance compared to Uni-IMU or Uni-Radar for all models. When $k=0.1$, in terms of single modality, the highest F1-score for eating gesture detection is 0.892 from Uni-Radar on MM-TCN-CMA, for drinking gesture detection is 0.906 from Uni-IMU on ResNet-BiLSTM. After fusing IMU and Radar data, the eating and drinking F1-scores improved to 0.922 and 0.920, respectively, using the MM-TCN-CMA approach. When applying a stricter threshold by increasing $k$ to 0.5, a general performance decline was observed across all models. Nevertheless, the fusion of IMU and radar data still outperformed uni-modality models, achieving eating and drinking F1-scores of 0.884 and 0.887, respectively. Furthermore, it was noted that drinking gesture detection performance using the Uni-IMU was generally higher than that of the Uni-Radar modality, a trend that became more pronounced when $k=0.5$.  A detailed individual performance analysis is presented in Fig.~\ref{indv_box}(a). Based on the Wilcoxon signed-rank test, for MM-TCN-CMA at $k=0.5$, the fusion of IMU and radar data significantly improved eating gesture detection compared to using Uni-IMU or Uni-Radar modalities ($p<0.001$). For drinking gesture detection, the fusion approach also demonstrated significantly higher performance than the Uni-Radar modality ($p<0.001$); however, no significant difference was observed when compared to the Uni-IMU modality.

\subsection{Modality-incomplete Experiments}
After validating the effectiveness of fusing IMU and radar data, we further designed experiments to assess the robustness of the proposed method under modality-incomplete conditions during the inference phase. Specifically, the models were trained on a modality-complete training set; however, during testing, one modality was intentionally dropped. For the missing radar experiment, the radar data were removed from all meal sessions in the test set, leaving only IMU data as input. Similarly, in the missing IMU experiment, only radar data were provided during testing. The sample-wise results are presented in Table~\ref{sample_missing_results}. It is observed that modality incompleteness leads to a reduction in performance. For the MM-TCN-CMA-based approach, the kappa scores for the missing radar and missing IMU experiments were 0.761 and 0.745, respectively, both lower than the modality-complete result of 0.777 (Table~\ref{sample_results}). However, when compared to the unimodal methods, the multimodal approach with missing radar still outperformed the Uni-IMU model (0.747), and the multimodal approach with missing IMU outperformed the Uni-Radar model (0.729). 

Table~\ref{missing_dam} further presents the segment-wise results. When $k=0.1$, eating F1-score under the missing IMU scenario (0.906) surpasses that of Uni-Radar (0.892), although lower than the IMU+Radar fusion (0.922). A similar trend is observed under the missing Radar condition. Notably, the drinking F1-score remains unaffected under the missing Radar scenario, which remained at 0.920. When $k=0.5$, under the missing Radar scenario, the eating and drinking F1-scores improve by 1.3\% and 3.2\%, respectively, over Uni-IMU. Specifically, the drinking performance is higher than IMU+Radar fusion (0.894 vs. 0.887). Similarly, under the missing IMU condition, eating and drinking F1-scores increase by 2.4\% and 4.1\%, respectively, compared to Uni-Radar. Fig.~\ref{indv_box}(b) shows that the detection performance of intake gestures using fusion with missing modalities is still higher than that achieved by Uni-IMU or Uni-Radar modalities with statistic significance ($p<0.05$), with the exception of eating gesture detection using Uni-IMU when $k=0.1$ ($p=0.08$).

\subsection{Ablation Study} \label{sec: abla}

\begin{table}[t]
	\caption{Segmental F1-scores ($k=0.5$) on different feature fusion methods}
	\label{fusion_comp}
	\begin{center}
		\scalebox{0.8}{
			\begin{tabular}{l|cc|cc|cc}
				\toprule
				\hline
				\multirow{2}*{Fusion Method}&\multicolumn{2}{c|}{IMU+Radar}&\multicolumn{2}{c|}{Missing Radar}&\multicolumn{2}{c}{Missing IMU}\\
				~&Eating&Drinking&Eating&Drinking&Eating&Drinking\\
				\midrule
				Add& 0.871 & 0.877  & 0.839& 0.890& 0.859 & 0.865 \\
				Contatenation & 0.874& 0.888  & 0.839 & 0.888 & 0.864 & 0.855\\
				Decision-level& 0.881& 0.886  & 0.846 & 0.891& 0.854 & 0.857\\
				CMA & 0.884& 0.887  & 0.845& 0.894& 0.865 & 0.879 \\
				
				\hline
				\bottomrule
		\end{tabular}}
	\end{center}
\end{table}

\begin{table}[t]
	\caption{Segmental F1-scores ($k=0.5$) on different training strategy}
	\label{train_comp}
	\begin{center}
		\scalebox{0.8}{
			\begin{tabular}{l|cc|cc|cc}
				\toprule
				\hline
				\multirow{2}*{Training Method}&\multicolumn{2}{c|}{IMU+Radar}&\multicolumn{2}{c|}{Missing Radar}&\multicolumn{2}{c}{Missing IMU}\\
				~&Eating&Drinking&Eating&Drinking&Eating&Drinking\\
				\midrule
				End-to-end& 0.864 & 0.885 & 0.828 & 0.872 & 0.839& 0.856 \\
				Two-step & 0.884& 0.887 & 0.845 & 0.894 & 0.865 & 0.879 \\
				\hline
				\bottomrule
		\end{tabular}}
	\end{center}
\end{table}

\begin{table}[t]
	\caption{Segment-wsie F1-scores results using MM-TCN-CMA model on radar and one-hand IMU}
	\label{one_hand_imu}
	\begin{center}
		\scalebox{0.8}{
			\begin{tabular}{l|cc|cc}
				\toprule
				\hline
				\multirow{2}*{Approach}&\multicolumn{2}{c|}{$k=0.1$}&\multicolumn{2}{c}{$k=0.5$}\\
				~&{Eating}&{Drinking}&{Eating}&{Drinking}\\
				\midrule
				Uni-IMU(one-hand)& 0.854 & 0.647 & 0.815 & 0.593 \\
				Uni-Radar & 0.892& 0.880 & 0.841 & 0.838  \\
				Fusion & 0.909& 0.898 & 0.871 & 0.861\\
				Fusion-missing IMU(one-hand) & 0.897& 0.890 & 0.851 & 0.853  \\
				Fusion-missing Radar & 0.864& 0.653 & 0.824 & 0.614\\
				\hline
				\bottomrule
		\end{tabular}}
	\end{center}
\end{table}

\begin{table}[t]
	\caption{Error analysis across different eating styles on MM-TCN-CMA approach at $k=0.5$}
	\label{type_error}
	\begin{center}
		\scalebox{0.8}{
			\begin{tabular}{l|lll|lll}
				\toprule
				\hline
				\multirow{2}*{Type}&\multicolumn{3}{c|}{\#FP}&\multicolumn{3}{c}{\#FN}\\
				~&Uni-IMU&Uni-Radar&Fusion&Uni-IMU&Uni-Radar&Fusion\\
				\midrule
				Fork \& Knife& 175 & 151 & 78 & 234 & 82 & 110  \\
				Spoon & 75 & 99 & 34 & 57 & 68 &56 \\
				Chopsticks &107& 108 & 65 & 109 & 242& 146\\
				Hand & 78& 115& 57& 140 &81 & 100 \\
				\hline
				\bottomrule	
		\end{tabular}}
	\end{center}
\end{table}

\begin{table}[t]
	\caption{Detailed eating gesture detection performance on MM-TCN-CMA approach}
	\label{det_per}
	\begin{center}
		\scalebox{0.8}{
			\begin{tabular}{l|ccc|ccc}
				\toprule
				\hline
				\multirow{2}*{Approach}&\multicolumn{3}{c|}{$k=0.1$}&\multicolumn{3}{c}{$k=0.5$}\\
				~&Precision&Recall&F1-score&Precision&Recall&F1-score\\
				\midrule
				Uni-IMU& 0.891 & 0.869 & 0.880 & 0.847 & 0.817& 0.832 \\
				Uni-Radar & 0.879& 0.905 & 0.892 & 0.841 & 0.841 & 0.841 \\
				Fusion & 0.942& 0.902 & 0.922 & 0.913 & 0.857 & 0.884\\
				Fusion-missing IMU & 0.915& 0.898 & 0.906 & 0.883 & 0.847& 0.865 \\
				Fusion-missing Radar & 0.917& 0.857 & 0.886 & 0.878 & 0.815& 0.845 \\
				\hline
				\bottomrule
		\end{tabular}}
	\end{center}
\end{table}

\subsubsection{Joint Operation}
At the feature fusion stage, various methods were investigated, including addition, concatenation, decision-level fusion, and cross-modal attention, as summarized in Table~\ref{fusion_comp}. Among these methods, the CMA approach achieved the highest performance.

\subsubsection{Training Strategy}
The proposed robust multimodal learning framework consists of several sub-modules, including MSFE, MAE, and FFM. In the MSFE module, the feature extraction encoders are pretrained and their weights are frozen during training, following a two-step training strategy. Alternatively, the entire framework can be trained from scratch in an end-to-end manner. In this experiment, both training setups were evaluated to assess their impact on performance, as shown in Table~\ref{train_comp}. The two-step training strategy yielded superior performance, supporting the hypothesis presented in Section~\ref{sec:tran}: joint training of all modules may lead to suboptimal results due to the risk of the undesirable scenario in which $\mathbb{F}^I$ adapts to $\mathbb{D}^{R2I}$.

\subsection{Radar + One-hand IMU Experiemnts} \label{sec: abla2}
The previous experiments assumed IMUs were worn on both wrists. However, in more practical scenarios, participants may only wear a single wrist-mounted IMU. This experiment investigates the performance of fusing radar with IMU exclusively from the dominant hand, defined as the hand with a higher frequency of eating gestures. The overall data processing pipeline remains unchanged, except that the IMU input dimension is reduced from 12 to 6. Importantly, ground truth labels are still based on both hands to ensure the detection of all intake gestures, rather than only those from the dominant hand. Table~\ref{one_hand_imu} presents the results. In the Uni-IMU setting, compared to the two-hand IMU configuration (Table~\ref{missing_dam}), the eating F1-score decreased by 1.7\% at $k = 0.5$, while the drinking F1-score exhibited a notable drop of 26.9\%. In the fusion mode, combining radar and one-hand IMU data yielded improvements in both eating and drinking F1-scores relative to the Uni-IMU and Uni-Radar baselines.

\begin{figure}[t]
	\centering
	\includegraphics[scale=0.25]{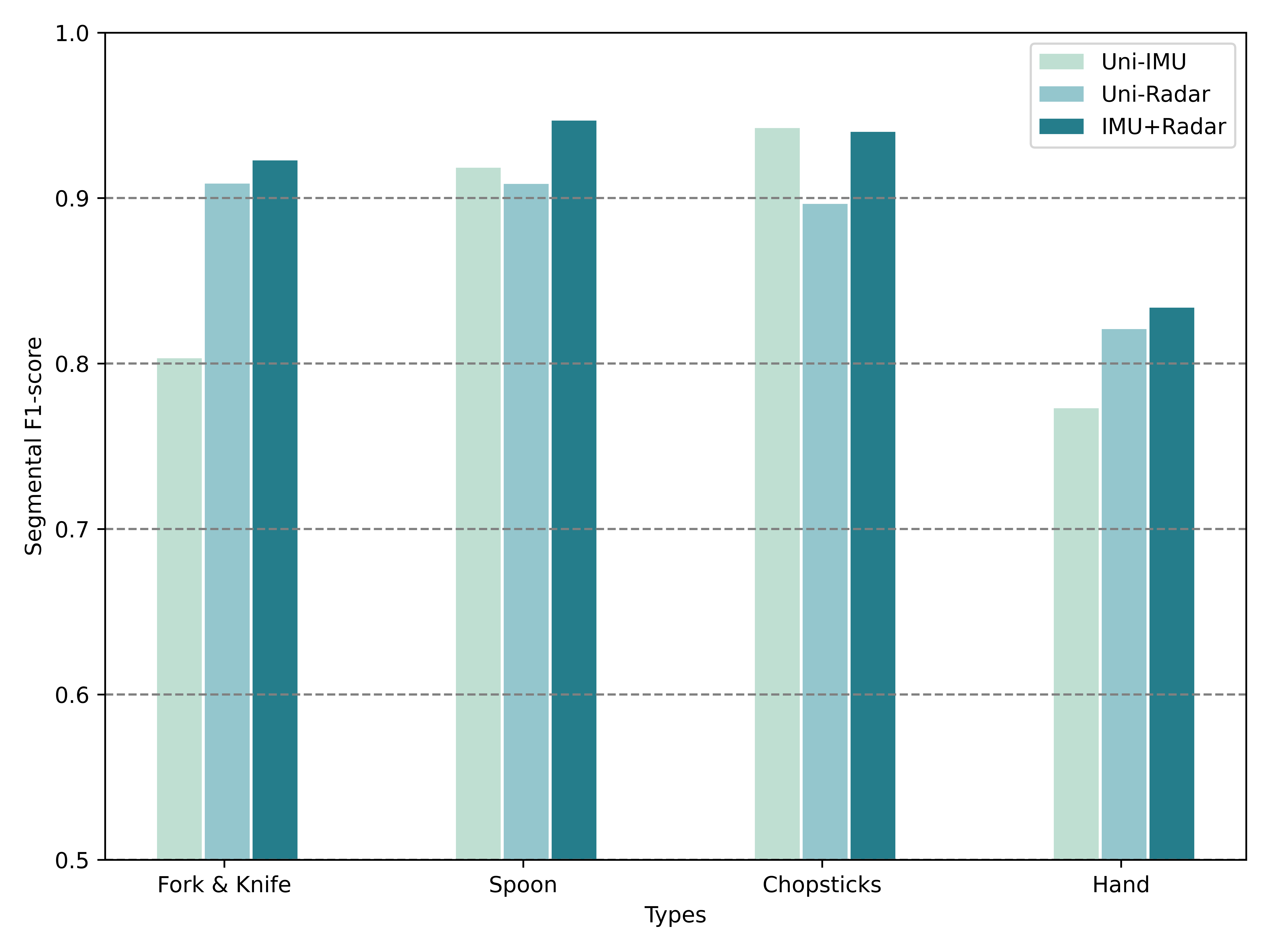}      
	\caption{Bar plots of the eating gesture detection performance across different eating styles, achieved by unimodal IMU-based, unimodal Radar-based, and multimodal fusion approaches using MM-TCN-CMA.}  
	\label{type_bar}
\end{figure}

\section{Discussion}
This study first investigated the efficacy of fusing radar and IMU sensors for hand-based intake gesture detection during meal sessions. As shown in Table~\ref{multi_results_2}, the combination of contactless radar and wearable IMU data achieved superior performance compared to using a single modality. The food intake dataset collected in this study contains four different eating styles: fork \& knife, spoon, chopsticks, and hand. Another objective was to investigate which eating style contributes most to the performance improvements. As illustrated in Fig. \ref{type_bar}, performance improvements were observed for the fork \& knife, spoon, and hand eating styles. However, for eating with chopsticks, the performance of the fusion method was comparable to that of the Uni-IMU approach. This suggests that the Uni-IMU-based method may have already reached optimal performance in detecting chopstick-based eating gestures, leaving limited room for further enhancement. Eating with hands yielded the lowest performance across all three approaches among the four eating styles, although an improvement was observed with the Radar-IMU-based approach. This finding implies that detecting eating gestures with hands is the most challenging task, consistent with the findings of our previous studies \cite{10684446,b9}. A detailed error analysis across different eating styles is presented in Table~\ref{type_error}. Compared to the number of false negatives (\#FN), the reduction in false positives (\#FP) was more pronounced across all eating styles when employing multimodal fusion. 

When comparing our work to prior studies utilizing multimodal sensors for intake gesture detection, the size of our dataset is comparable to those used in similar research. In these studies, different sensors are typically integrated into a wearable system , or an ambient system. Only one study \cite{b15} investigates the combination of data from wearable IMU and ambient-based camera devices. When comparing our work to prior studies on Radar-IMU combined HAR \cite{b19,b20}, our dataset size is significantly larger. Moreover, the experimental scenario in our study involves continuous meal sessions.

\begin{figure}[t]
	\centering
	\includegraphics[scale=0.3]{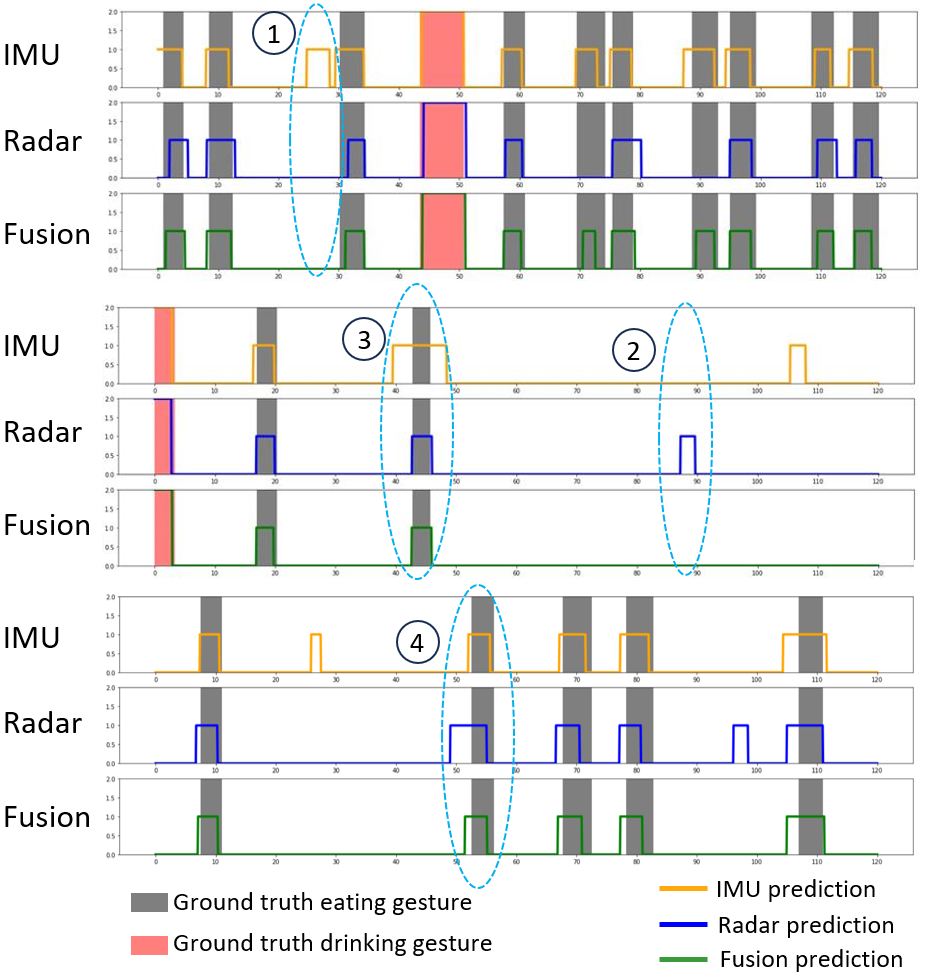}      
	\caption{Examples of predictions generated by unimodal IMU-based, unimodal Radar-based, and multimodal fusion approaches. The circled cases (1)(2) represent improvements in detection when fusing IMU+Radar, and circled cases (3)(4) illustrate enhancements in temporal segmentation precision for detected gestures.}  
	\label{exam_pre}
\end{figure}

This study also addressed the missing modality issue, a challenging and meaningful research problem in HAR, within the context of robust multimodal learning that involves heterogeneous wearable and ambient modalities. By leveraging modality adaption, the model achieves higher performance when tested with incomplete modalities compared to a unimodal approach. This suggests that the MAE can provide additional complementary features from the perspective of the target modality when processing the source modality.

Currently, wearable and ambient sensors are two mainstream approaches for automated intake gesture detection. Specifically, IMU sensors are commonly used in wearable devices, while radar sensors are emerging as privacy-preserving ambient sensors. Exploring the potential of fusing these two modalities is worthwhile, as we believe this approach can extend beyond intake gesture detection to broader HAR tasks, such as those in elderly care, smart homes, and driving environments. 

An analysis of the detailed performance results presented in Table \ref{det_per} reveals that the primary improvement stems from a reduction in the number of FPs, which in turn leads to an increase in precision. This finding suggests that the proposed multimodal framework effectively leverages the complementary information from IMU and radar sensors, helping to suppress ambiguous or noisy inputs that would otherwise result in false positives when relying on a single modality. Examples of this effect are illustrated in Fig. \ref{exam_pre}(1)(2). Additionally, it is observed that as the IoU threshold $k$ increases from 0.1 to 0.5, the decreases in precision and recall for the multimodal framework (2.9\% and 4.5\%) are smaller compared to those for the unimodal baselines (IMU: 4.4\% and 5.2\%, Radar: 3.8\% and 6.4\%). These results suggest that the integration of radar motion profiles and IMU body motion signals enables the model to produce more accurate and tighter gesture boundaries, resulting in higher overlap ratios, as demonstrated in Fig. \ref{exam_pre}(3)(4). Overall, these observations indicate that the proposed multimodal framework not only enhances classification performance but also improves segmentation quality. Furthermore, these benefits are maintained even when one modality is absent during the inference phase, underscoring the framework's robustness to partial sensor availability.

Despite the contributions of this research, there are still some limitations. First, the training phase relies on modality-complete data, which inherently restricts the size of the available training set. Future work could focus on developing models that are capable of training with modality-incomplete data, thus allowing for the use of larger and more diverse datasets. Secondly, the integration of the modality adaptation module introduces additional parameters, which increases the computational overhead during both the training and inference phases. Third, the current study focuses primarily on a single-person scenario, which simplifies the problem setting. In more complex situations involving multiple individuals equipped with IMUs within the radar’s field of view, it becomes necessary to associate radar signal with the corresponding participant’s IMU data prior to sensor fusion. Addressing these challenges would enhance the scalability and efficiency of the proposed framework.

\section{Conclusion}
In this study, we propose a robust multimodal learning framework for intake gesture detection, utilizing contactless radar and wearable IMU sensors, with the capability to process data with missing modalities. The results obtained from the collected dataset validate the efficacy of the model. Specifically, the fusion of wearable and ambient sensor signals enhances the performance of intake gesture detection. Additionally, the multimodal model demonstrates robustness when either radar or IMU data is missing during the inference phase. The proposed framework has significant potential for application in other HAR studies.

\section*{Acknowledgment}
The authors would like to thank the participants who participated in the experiments for their efforts and time. The computational resources and services used in this work were provided by the VSC (Flemish Supercomputer Center), funded by the Research Foundation Flanders (FWO) and the Flemish Government – department EWI.
% BibTex
\bibliographystyle{IEEEtran}
\bibliography{papers}
\end{document}